\pdfoutput=1

\documentclass[11pt]{article}

\usepackage{acl}

\usepackage{times}
\usepackage{latexsym}

\usepackage[T1]{fontenc}

\usepackage[utf8]{inputenc}

\usepackage{microtype}

\usepackage{booktabs}       
\usepackage{amsfonts}       
\usepackage{nicefrac}       
\usepackage{microtype}      
\usepackage{xcolor}         
\usepackage{latexsym}
\usepackage{url}
\usepackage{comment}
\usepackage{amsmath,amssymb}

\usepackage{todonotes}
\usepackage{booktabs}
\usepackage{footnote}
\usepackage{microtype}
\usepackage{graphicx,paralist,multirow,makecell}
\usepackage{booktabs}
\usepackage{color}
\usepackage{multicol}
\usepackage{subcaption}
\usepackage{xcolor,colortbl}
\usepackage{wrapfig}
\usepackage[export]{adjustbox}
\usepackage{tablefootnote}
\usepackage{diagbox}
\usepackage{slashbox}

\newcommand{\tonics}{\texttt{TOnICS}}

\usepackage{pifont}
\newcommand{\cmark}{\ding{51}}%
\newcommand{\xmark}{\ding{55}}%

\title{Curriculum Learning for Data-Efficient Vision-Language Alignment}

\author{Tejas Srinivasan \phantom{\and}
  Xiang Ren \phantom{\and}
  Jesse Thomason\\
  University of Southern California\\
  \texttt{tejas.srinivasan@usc.edu}}

\begin{document}
\maketitle


\begin{abstract}
    
Aligning image and text encoders from scratch using contrastive learning requires large amounts of paired image-text data. 
We alleviate this need by aligning individually pre-trained language and vision representation models using a much smaller amount of paired data, augmented with a curriculum learning algorithm to learn fine-grained vision-language alignments. 
\tonics\ (\textbf{T}raining with \textbf{On}tology-\textbf{I}nformed \textbf{C}ontrastive \textbf{S}ampling) initially samples minibatches whose image-text pairs contain a wide variety of objects to learn object-level alignment, and progressively samples minibatches where all image-text pairs contain the same object to learn finer-grained contextual alignment. 
Aligning pre-trained BERT and VinVL models to each other using \tonics\ outperforms CLIP on downstream zero-shot image retrieval while using less than 1\% as much training data.
\end{abstract}

\section{Introduction}
\label{sec:intro}

Aligned representations for language and vision---which encode texts and corresponding images in a common latent space---are necessary to perform effective cross-modal retrieval. 
CLIP~\cite{radford2021learning} and ALIGN~\cite{jia2021scaling} train individual text and image encoders from scratch to produce aligned image-text representations. Their encoders demonstrate strong cross-modal alignment, evidenced by strong performance on zero-shot retrieval tasks. However, these models were trained on proprietary datasets of 400M and 1B image-text pairs respectively, on hundreds of GPUs and TPUs, which is infeasible for non-industry practitioners. 

CLIP and ALIGN align their encoders using the contrastive InfoNCE objective~\cite{oord2018representation}, which seeks to maximize the mutual information between image and text representations. 
In the InfoNCE objective, the model must correctly identify the positive image-text pair from among a set of negatives formed by the other minibatch pairs.

\begin{figure}[t]
    \centering
    \includegraphics[width=\columnwidth]{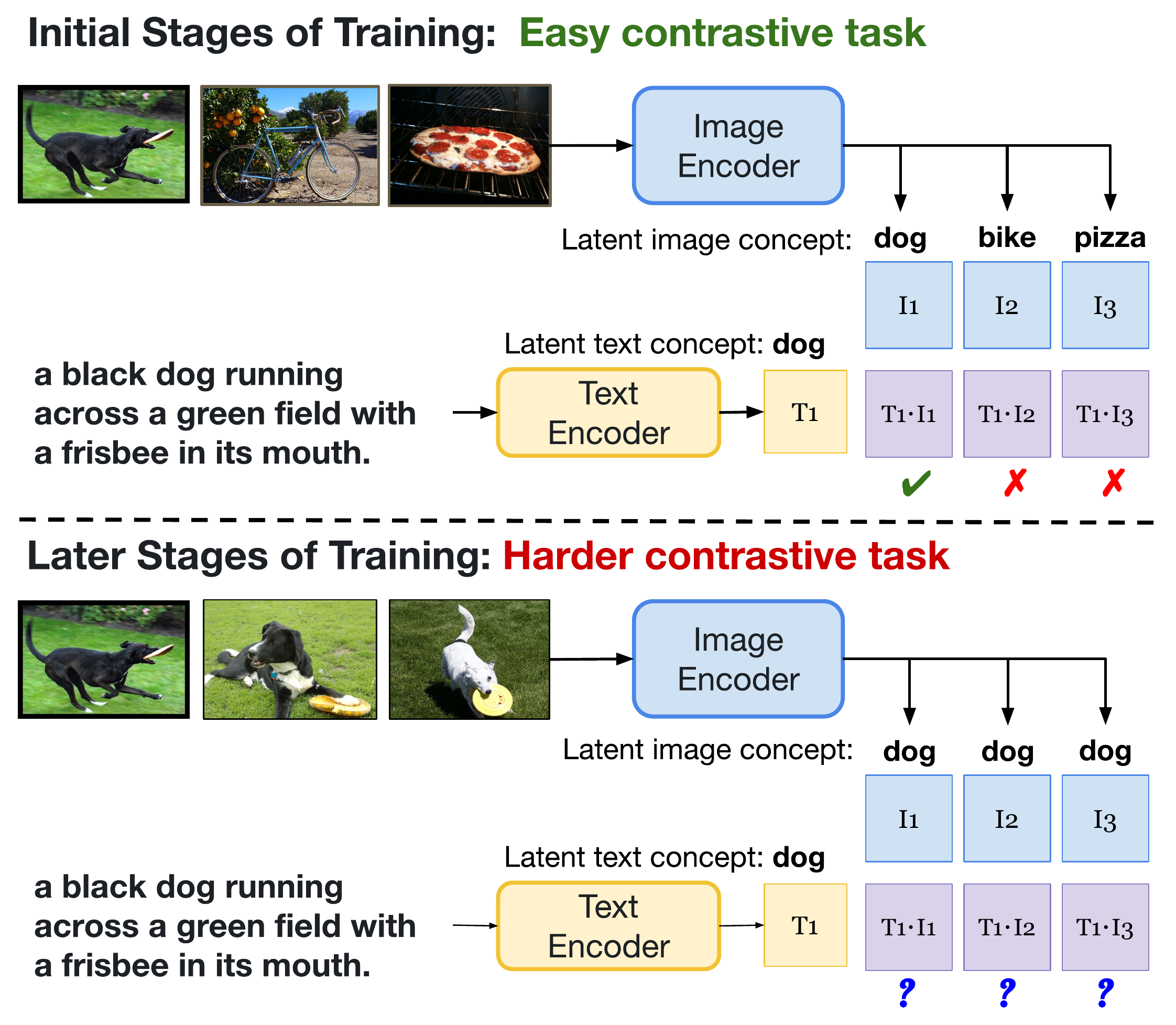}
    \caption{We propose \tonics, a curriculum learning algorithm for contrastive alignment of language and vision encoders.}
    \label{fig:page1}
\end{figure}

Since samples within a minibatch act as negative samples for each other in the InfoNCE objective, the minibatch determines the granularity of alignment that is learned. 
Minibatches constructed by random sampling contain a large variety of objects in the images and texts (Figure~\ref{fig:page1}, top). 
To correctly match a \textit{dog}-related caption to its image, it is sufficient to identify that the retrieved image must contain a dog, since the vast majority of randomly sampled negative images will not contain a dog. 
Thus, random minibatch sampling reduces the contrastive task to object-matching, for which object-level vision-language alignment suffices.

When minibatches are sampled such that the images contain the same objects, object-level alignments no longer suffice (Figure~\ref{fig:page1}, bottom). 
The contrastive task can no longer be solved by identifying that the retrieved image must contain a dog, since all the negative images will also have a dog. 
The model must produce language and vision representations that encode shared \textit{context}-level information, resulting in a finer-grained alignment.

In this work, rather than training our image and text encoders from scratch, we leverage rich single-modality pre-trained models---BERT~\cite{devlin2018bert} for language, VinVL~\cite{zhang2021vinvl}\footnote{We use VinVL to refer to their pre-trained object detector.} for vision---and align them to each other using the InfoNCE contrastive objective. 
We perform the vision-language alignment using \tonics, a novel ontology-based curriculum learning algorithm.
\tonics\ initiates training with an easy contrastive task by sampling minibatches randomly and progressively makes the contrastive task harder by constructing minibatches containing the same object class in the image and text inputs.
We show that our learned representations have strong cross-modal alignment---outperforming CLIP on zero-shot Flickr30K image retrieval---while using less than 1\% as much paired image-text training data.

\section{Contrastive Vision-Language Alignment}
\label{sec:training-steup}
We align language representations from BERT~\cite{devlin2018bert} and visual representations from a VinVL object detector~\cite{zhang2021vinvl}. Our BERT-VinVL Aligner model is similar to the phrase grounding model from~\citet{gupta2020contrastive}.

At every training step, the input to the model is a minibatch of $N_B$ triplets, where each triplet $X_i = \{ t^i, v^i, w\}$ comes from an image-text pair. 
Each image caption $t^i$ is encoded using BERT. 
The caption contains a noun $w$, whose word representation is denoted as $h^i$.
For the corresponding image, $v^i$ is a set of region features extracted from a frozen pre-trained VinVL object detector.\footnote{Region features provided at \url{https://github.com/pzzhang/VinVL/blob/main/DOWNLOAD.md}} We add a learnable linear projection atop these region features.

\begin{figure}[t]
    \centering
    \includegraphics[width=\columnwidth]{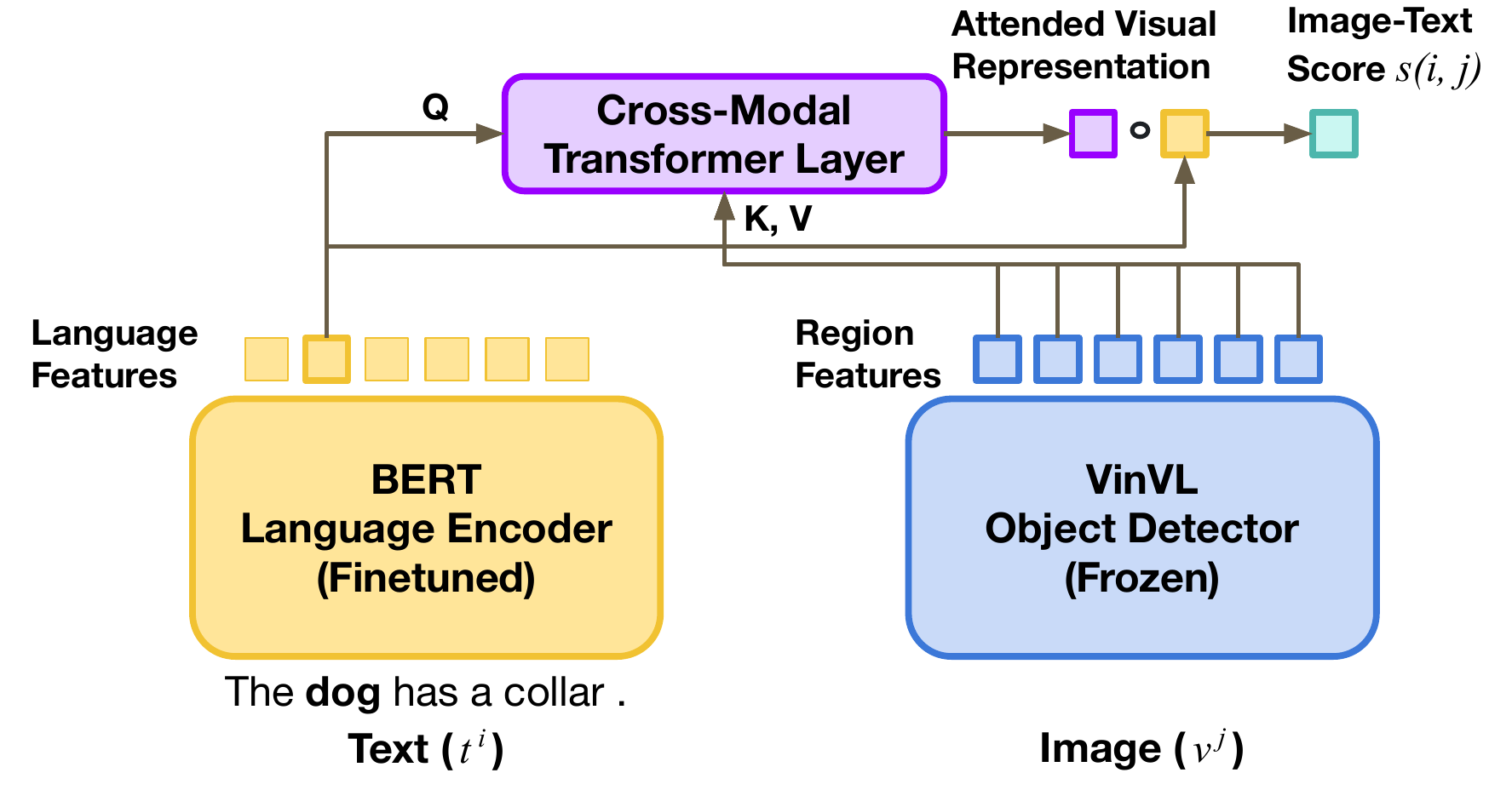}
    \caption{Our BERT-VinVL Aligner model scores every image-text combination ($t^i, v^j$) in the minibatch.}
    \label{fig:model}
\end{figure}

In the cross-modal interaction, we employ a single Transformer~\cite{vaswani2017attention} layer that uses $i$-th noun representation $h^i$ as the query and $j$-th image features $v^i$ as the keys and values. 
This layer outputs a visual representation $v_{att}(i, j)$, which is an attended representation of the $j$-th image, conditioned on the noun from the $i$-th caption. We then compute a dot product between the $i$-th noun representation $h^i$ and the attended representation of $j$-th image $v_{att}(i, j)$ to get an image-text score $s(i, j) = \phi(h^i, v_{att}(i,j))$ (Figure~\ref{fig:model}).

To align the noun representation $h^i$ to its corresponding image $v^i$, we use the InfoNCE loss~\cite{oord2018representation} which maximizes a lower bound of the mutual information between $h^i$ and $v_{att}(i,i)$. 
InfoNCE minimizes the cross-entropy of correctly retrieving an image $v^i$ from the set of all minibatch images, given the query noun representation $h^i$, with other instances in the minibatch acting as negative samples. 
We refer to the objective in this setup as the image retrieval loss, $\mathcal{L}_{IR}$:
\begin{align*}
    \mathcal{L}_{IR}(i) = - \log \frac{\exp(s(i,i))}{\sum_{j=1}^{N_B} \exp(s(i, j))}
\end{align*}
The training loss $\mathcal{L}_{IR}$ is the mean loss $\mathcal{L}_{IR}(i)$ over all images $i=\{1...N_B\}$ in the minibatch $\mathcal{B}$. We also similarly define a text retrieval loss, $\mathcal{L}_{TR}$, where the image $v^i$ is used to retrieve the correct noun representation $h^i$:
\begin{align*}
    \mathcal{L}_{TR}(i) = - \log \frac{\exp(s(i, i))}{\sum_{j=1}^{N_B} \exp(s(j, i))}
\end{align*}

We experiment with training our model using just the image retrieval loss $\mathcal{L}_{IR}$, as well as the sum of the two losses $\mathcal{L}_{IR} + \mathcal{L}_{TR}$.

\section{\tonics: Training with Ontology Informed Contrastive Sampling}
\label{sec:tonics}
\begin{figure*}
    \centering
        \includegraphics[width=\textwidth]{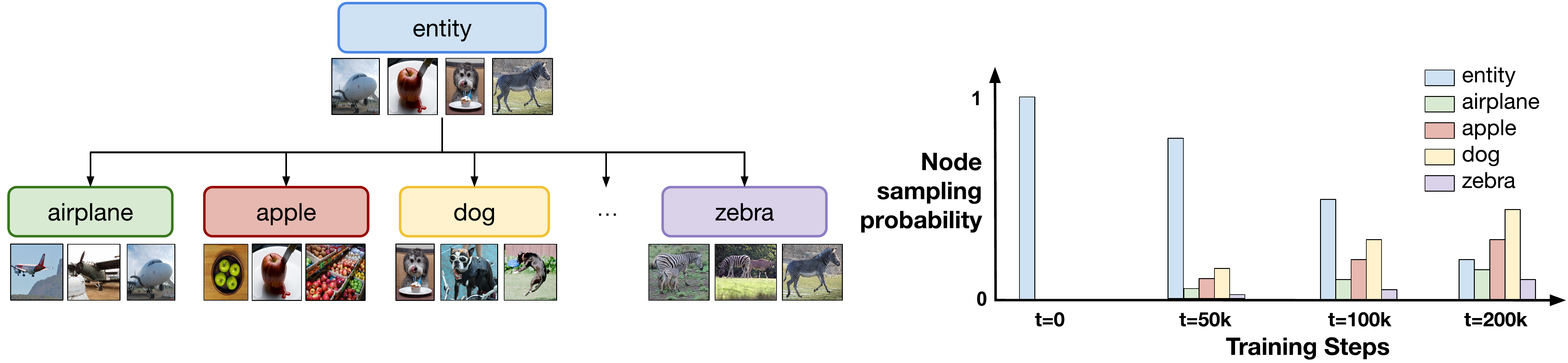}
    \caption{\tonics\ selects image-text pairs for the minibatch by first sampling a node $\eta$ from an ontology, according to a distribution $P_S(\eta)$.
    Sampling the root \textit{entity} node yields easy minibatches containing pairs with a variety of objects, whereas sampling one of its children \textit{object nodes} yields harder minibatches containing pairs sharing a common object, such as \textit{apple} or \textit{dog}, in a variety of contexts (left). \tonics\ performs curriculum learning by moving node sampling mass away from the root entity node to the object nodes as training progresses (right).}
    \label{fig:tonics}
\end{figure*}

As noted above, negative samples for the contrastive learning  objective come from other pairs in the minibatch. Therefore, the minibatch sampling itself influences the alignment learned by the model. We hypothesize that sampling minibatches randomly will yield object-level alignments, while sampling harder minibatches containing the same object in the image may result in fine-grained contextual alignments.

We introduce \tonics, \textbf{T}raining with \textbf{On}tology-\textbf{I}nformed \textbf{C}ontrastive \textbf{S}ampling (Figure~\ref{fig:tonics}), a curriculum learning algorithm that initially seeks to align vision and language representations at the object level, and later learns contextual alignments. \tonics\ initiates the training by generating minibatches with randomly sampled image-text-noun triplets. As training progresses, \tonics\ samples harder minibatches whose instances share the same object class in the image. 

\paragraph{Ontology Construction} We begin by extracting object detections from our training images using the pre-trained VinVL model. 
We next map each noun in the training data to an object class, wherever possible, resulting in a set of object classes $\Theta$.
Every object class $o \in \Theta$ has a corresponding set of nouns $w(o)$. 
For instance, the object class \textit{dog}'s noun set $w(o)=\{$\textit{dog, dogs, puppy}$\}$. 

We construct the ontology (Figure~\ref{fig:tonics}, left), which contains an \textit{entity} root node and its children \textit{object nodes} $\eta_o$, each corresponding to an object class $o$.
Every object node $\eta_o$ has a corresponding set of triplet instances $X(\eta_o)$, a subset of the full training dataset whose triplet instances all contain the same object class $o$ in the image, and all containing a noun from the noun set $w(o)$ in the caption.

\paragraph{\tonics\ Minibatch Sampling} At every training step, \tonics\ proceeds in two stages. First, a node $\eta$ is sampled from the ontology, according to a sampling probability distribution $P_S(\eta)$. 
Second, we sample a minibatch according to the node that was just sampled. 
If we sample the entity node $\eta_e$, we sample the minibatch by sampling $N_B$ instances from the full training data at random. If we sample an object node $\eta_o$, we sample $N_B$ instances from the corresponding set $X(\eta_o)$, ensuring the minibatch contains images with the same object.

\begin{table*}[t]
\small
    \centering
    \begin{tabular}{l l l l c c c c c c c c}
    \toprule
         & & \multirow{3}{*}{\makecell[l]{Minibatch\\Sampling\\Method}}& & \multicolumn{4}{c}{Zero-Shot Flickr30K} & \multicolumn{4}{c}{MS-COCO} \\
         & \multirow{2}{*}{\makecell[l]{\# Image-\\Text Pairs}} & & & \multicolumn{2}{c}{Image Retrieval} & \multicolumn{2}{c}{Text Retrieval} & \multicolumn{2}{c}{Image Retrieval} & \multicolumn{2}{c}{Text Retrieval} \\
         Model & & & $\mathcal{L}_{TR}$ & R@1 & R@5 & R@1 & R@5 & R@1 & R@5 & R@1 & R@5 \\
    \midrule 
    CLIP-ViT-B/32 & 400M & Random & - & 58.66 &	83.38 & \textbf{79.2} & \textbf{95} & 30.45 &	56.02 & 50.12 &	75.02 \\
    \midrule
    \multirow{4}{*}{\makecell[l]{BERT-VinVL\\Aligner}} & 2.84M& Random & \xmark & 58.18 &	84.24 & 22.2 & 47.9 & 42.67 &	74.43 & 15.5 & 37.7 \\
    \vspace{0.3em}
    & 2.84M & \tonics & \xmark & \textbf{60.32} & 85.14 & 24.4 & 49 & 47.94 & 77.38 & 16.1 & 35.1  \\
    & 2.84M & Random & \cmark & 58.9 &	84.6 & 76.1 & 93.3 & 42.74 &	74.37 & 59.84 &	86.46 \\
    & 2.84M & \tonics & \cmark & 59.7 & \textbf{85.24} & 76.6 & 94.1 & \textbf{48.26} & \textbf{77.87} & \textbf{65.44} & \textbf{89.36}  \\
    \bottomrule
    \end{tabular}
    \caption{Results of our BERT-VinVL Aligner model on image and text retrieval, compared to a CLIP model.
    Numbers in bold represent the best results among our model and CLIP.}
    \label{tab:results}
\end{table*}

\paragraph{\tonics\ Curriculum Refresh} The curriculum is formed by varying the nodes' sampling probability distribution throughout training. 
We initialize training by setting $P_S(\eta_e)=1$ and $P_S(\eta_o) = 0$ for all object nodes. 
After every fixed number of training steps, we evaluate the model's image retrieval performance on a set of 100 held-out instances. 
If the held-out retrieval accuracy is greater than a certain threshold, we say that the model has learned the object-level alignment task, and we can start introducing harder minibatches in the training by \textit{refreshing} the curriculum. 
The refresh step is performed by multiplying the entity node's current sampling probability $P_S(\eta_e)$ by a factor $\alpha; \alpha<1$. 
The remaining probability mass $(1-\alpha) \times P_S(\eta_e)$ is distributed among the object nodes. For each object node $\eta_o$, we update its sampling probability:
\begin{align*}
    P_S(\eta_o) = P_S(\eta_o) + (1-\alpha) P_S(\eta_e) \times \frac{|X(\eta_o)|}{\sum |X(\eta_o)|}.
\end{align*}
Object classes that are more common in the training data have more sampling probability mass distributed to their object node $\eta_o$, by weighting mass according to the size of the node's instance set, $|X(\eta_o)|$. 
With each curriculum refresh, sampling mass is pushed down from the entity node to the object nodes, as long as $P_S(\eta_e)$ does not fall below a fixed threshold $\beta$. 
Thresholding $P_S(\eta_e)$ ensures the model still sees random minibatches and does not forget the initially learned object-level alignments.

\section{Experiment Details}
\label{sec:exp-details}
We train our BERT-VinVL model on MS-COCO and Conceptual Captions. 
We compare our model against CLIP on downstream retrieval tasks.

\subsection{Training Data and Ontology}
\label{subsec:data}
We train our model on image-text pairs from a combination of MS-COCO~\cite{chen2015microsoft} and Conceptual Captions~\cite{Sharma2018}. 
Our triplet instances only contain nouns which we wish to explicitly align with the visual modality. 
Each noun in the training data is initially mapped to the object class with maximum noun-object PMI, calculated over training pairs with object detections, and then adjusted by hand to correct erroneous mappings. 
Object classes containing fewer than 5000 instances in the training dataset are filtered out. 
This finally results in a set of 406 nouns, each noun corresponding to one of the 244 object categories $\Theta$. 
For every image-text pair in the original training dataset, we create one triplet for each noun in our set of 406 nouns that the text contains. 

Our final training data consists of 5.8M triplet instances corresponding to 2.84M image-text pairs (2.26M from Conceptual Captions, 580K from MS-COCO) from 2.4M unique images. 
The ontology for \tonics\ is constructed by creating an object node for each of the 244 object categories, which are children of the root \textit{entity} node.

\subsection{Implementation Details}

We use pre-trained BERT-base as our text encoder. For our image encoder, we use VinVL, a pre-trained object detector that detects regions of interest (ROIs) in the image and outputs pooled CNN features for all ROIs. 
We use pre-extracted ROI features and treat the VinVL encoder as frozen, as we cannot backpropagate through the object detector.

All our models are trained for 500K iterations with a batch size of $N_B=256$, yielding 255 negative pairs for every positive pair. Each model was trained on a single V100 GPU for 6 days, compared to CLIP which used 256 V100 GPUs for 12 days.

After every 5K iterations, we evaluate retrieval over a set of held-out instances and perform a curriculum refresh step if the held-out accuracy is at least 90\%. 
When performing a refresh step, we retain $\alpha=90\%$ of \textit{entity}'s sampling probability, so long as the probability does not fall below $\beta=0.2$.

\subsection{Baselines and Evaluation}

To compare the effect of using pre-trained unimodal encoders at the start of the alignment process, we compare our model against CLIP~\cite{radford2021learning}. 
CLIP also uses separate image and text encoders, aligned using a contrastive loss with image-text data. 
Unlike our BERT-VinVL Aligner model, CLIP trains the two encoders from scratch, and uses significantly more  paired image-text data---400M pairs, compared to our 2.84M pairs. 
Since we use the base variant of BERT, we compare against the CLIP-ViT-B/32 variant.\footnote{Checkpoint provided at \url{https://huggingface.co/openai}} 
We do not compare against ALIGN as they have not released their base model checkpoint.

To evaluate the utility of our \tonics\ algorithm, we also train our BERT-VinVL Aligner using a \textbf{Random} minibatch sampling baseline, where the minibatch instances are always randomly sampled throughout the training process.

We directly evaluate our trained Aligner model's (as well as pre-trained CLIP) on image and text retrieval. 
Specifically, we perform zero-shot retrieval on the test set of Flickr30K~\cite{plummer2015flickr30k}, which contains 1,000 images. 
We also perform retrieval evaluation on the MS-COCO test set, which contains 5,000 images.
This latter evaluation is not zero-shot since our training data contains images from the MS-COCO train set. 
We compare the Recall@1 and Recall@5 of all models.

\section{Results and Discussion}
\label{sec:results}
In Table~\ref{tab:results}, we directly transfer both our trained BERT-VinVL Aligner model and pre-trained CLIP to the downstream task of image and text retrieval. 
Since our models are trained using retrieval objectives, we perform the retrieval evaluation using the same setup as training.

The Flickr30K evaluation is zero-shot for both CLIP and our BERT-VinVL Aligner model since neither model's training data contains images from the Flickr30K train set. 
We see that even with the Random minibatch sampling and only the image retrieval loss, $\mathcal{L}_{IR}$, our BERT-VinVL Aligner achieves approximately the same image retrieval performance as CLIP. 
When the Aligner is trained with our \tonics\ curriculum learning algorithm, we get a 1.5\% improvement on R@1 over CLIP.

However, this model fails to do well at the text retrieval task. 
Adding the text retrieval loss $\mathcal{L}_{TR}$ leads to substantial improvements in downstream text retrieval, with the Random baseline performing only 3\% worse than CLIP. 
We further see that training with \tonics\ leads to only slight improvements in Flickr30K text retrieval. 
Adding the text retrieval loss slightly hurts image retrieval performance, but still does better than CLIP by 1\%.

Since our model, unlike CLIP, includes MS-COCO training images in the training data, it significantly outperforms CLIP on the MS-COCO retrieval evaluation. 
Hence, we compare our \tonics\ algorithm to the Random baseline on the MS-COCO evaluation. We see that \tonics\ leads to significant improvements in image retrieval (> 5\%), both when the text contrastive loss is and isn't used. 
We once again see that the text retrieval performance is very poor without the text retrieval objective during training, but improves significantly with it. 
\tonics\ results in a 5\% improvement over the Random baseline in text retrieval as well.

Minibatch sampling with \tonics\ results in large gains in in-distribution retrieval evaluation (MS-COCO) as well as small improvements in zero-shot retrieval (Flickr30K). Training BERT-VinVL with \tonics\ yields better zero-shot image retrieval performance than CLIP, even with substantially less training data.

\section{Conclusions and Future Work}
In this work, we align individually pre-trained language and vision encoders---BERT and VinVL, respectively---using a novel curriculum learning algorithm called \tonics. 
Our aligned model is able to achieve better downstream zero-shot image retrieval performance than CLIP, in spite of being trained with less than 1\% as many image-text training pairs. 
We further show that our \tonics\ algorithm leads to gains in both in-domain and zero-shot retrieval tasks.

\bibliography{anthology,custom}

\begin{thebibliography}{10}
\expandafter\ifx\csname natexlab\endcsname\relax\def\natexlab#1{#1}\fi

\bibitem[{Chen et~al.(2015)Chen, Fang, Lin, Vedantam, Gupta, Doll{\'a}r, and
  Zitnick}]{chen2015microsoft}
Xinlei Chen, Hao Fang, Tsung-Yi Lin, Ramakrishna Vedantam, Saurabh Gupta, Piotr
  Doll{\'a}r, and C~Lawrence Zitnick. 2015.
\newblock {Microsoft COCO Captions}: Data collection and evaluation server.
\newblock \emph{arXiv preprint arXiv:1504.00325}.

\bibitem[{Devlin et~al.(2019)Devlin, Chang, Lee, and
  Toutanova}]{devlin2018bert}
Jacob Devlin, Ming-Wei Chang, Kenton Lee, and Kristina Toutanova. 2019.
\newblock {BERT}: Pre-training of deep bidirectional transformers for language
  understanding.
\newblock In \emph{North {A}merican Chapter of the Association for
  Computational Linguistics (NAACL)}.

\bibitem[{Gupta et~al.(2020)Gupta, Vahdat, Chechik, Yang, Kautz, and
  Hoiem}]{gupta2020contrastive}
Tanmay Gupta, Arash Vahdat, Gal Chechik, Xiaodong Yang, Jan Kautz, and Derek
  Hoiem. 2020.
\newblock Contrastive learning for weakly supervised phrase grounding.
\newblock In \emph{European Conference on Computer Vision (ECCV)}.

\bibitem[{Jia et~al.(2021)Jia, Yang, Xia, Chen, Parekh, Pham, Le, Sung, Li, and
  Duerig}]{jia2021scaling}
Chao Jia, Yinfei Yang, Ye~Xia, Yi-Ting Chen, Zarana Parekh, Hieu Pham, Quoc Le,
  Yun-Hsuan Sung, Zhen Li, and Tom Duerig. 2021.
\newblock Scaling up visual and vision-language representation learning with
  noisy text supervision.
\newblock In \emph{International Conference on Machine Learning (ICML)}.

\bibitem[{Oord et~al.(2018)Oord, Li, and Vinyals}]{oord2018representation}
Aaron van~den Oord, Yazhe Li, and Oriol Vinyals. 2018.
\newblock Representation learning with contrastive predictive coding.
\newblock \emph{arXiv preprint arXiv:1807.03748}.

\bibitem[{Plummer et~al.(2015)Plummer, Wang, Cervantes, Caicedo, Hockenmaier,
  and Lazebnik}]{plummer2015flickr30k}
Bryan~A Plummer, Liwei Wang, Chris~M Cervantes, Juan~C Caicedo, Julia
  Hockenmaier, and Svetlana Lazebnik. 2015.
\newblock {Flickr30k Entities}: Collecting region-to-phrase correspondences for
  richer image-to-sentence models.
\newblock In \emph{International Conference on Computer Vision (ICCV)}.

\bibitem[{Radford et~al.(2021)Radford, Kim, Hallacy, Ramesh, Goh, Agarwal,
  Sastry, Askell, Mishkin, Clark et~al.}]{radford2021learning}
Alec Radford, Jong~Wook Kim, Chris Hallacy, Aditya Ramesh, Gabriel Goh,
  Sandhini Agarwal, Girish Sastry, Amanda Askell, Pamela Mishkin, Jack Clark,
  et~al. 2021.
\newblock Learning transferable visual models from natural language
  supervision.
\newblock \emph{arXiv preprint arXiv:2103.00020}.

\bibitem[{Sharma et~al.(2018)Sharma, Ding, Goodman, and Soricut}]{Sharma2018}
Piyush Sharma, Nan Ding, Sebastian Goodman, and Radu Soricut. 2018.
\newblock {Conceptual Captions}: A cleaned, hypernymed, image alt-text dataset
  for automatic image captioning.
\newblock In \emph{Association for Computational Linguistics (ACL)}.

\bibitem[{Vaswani et~al.(2017)Vaswani, Shazeer, Parmar, Uszkoreit, Jones,
  Gomez, Kaiser, and Polosukhin}]{vaswani2017attention}
Ashish Vaswani, Noam Shazeer, Niki Parmar, Jakob Uszkoreit, Llion Jones,
  Aidan~N Gomez, {\L}ukasz Kaiser, and Illia Polosukhin. 2017.
\newblock Attention is all you need.
\newblock In \emph{Neural Information Processing Systems (NeurIPS)}.

\bibitem[{Zhang et~al.(2021)Zhang, Li, Hu, Yang, Zhang, Wang, Choi, and
  Gao}]{zhang2021vinvl}
Pengchuan Zhang, Xiujun Li, Xiaowei Hu, Jianwei Yang, Lei Zhang, Lijuan Wang,
  Yejin Choi, and Jianfeng Gao. 2021.
\newblock {VinVL}: Revisiting visual representations in vision-language models.
\newblock In \emph{Computer Vision and Pattern Recognition (CVPR)}.

\end{thebibliography}
\bibliographystyle{acl_natbib}

\appendix

\end{document}